\documentclass[preprint,5p, times, twocolumn]{elsarticle}

\usepackage{amssymb}
\usepackage{amsmath}
\usepackage{tikz}
\usepackage[hyphens]{url}

\usepackage{xcolor}
\definecolor{light-gray}{gray}{0.95}
\newcommand{\code}[1]{\colorbox{light-gray}{\texttt{#1}}}

\journal{Future Generation Computer Systems}

\begin{document}

\begin{frontmatter}



\title{Concurrent Vertical and Horizontal Federated Learning with Fuzzy Cognitive Maps}

\author[cunef,chile]{Jose L. Salmeron}
\ead{joseluis.salmeron@cunef.edu}
\address[cunef]{CUNEF Universidad, Madrid, Spain}
\address[chile]{Universidad Autonoma de Chile, Chile}

\author[cunef]{Irina Ar\'evalo\corref{cor1}}
\ead{irina.arevalo@cunef.edu}

\cortext[cor1]{Corresponding author}

\begin{abstract}
Data privacy is a major concern in industries such as healthcare or finance. The requirement to safeguard privacy is essential to prevent data breaches and misuse, which can have severe consequences for individuals and organisations. Federated learning is a distributed machine learning approach where multiple participants collaboratively train a model without compromising the privacy of their data. However, a significant challenge arises from the differences in feature spaces among participants, known as non-IID data. This research introduces a novel federated learning framework employing fuzzy cognitive maps, designed to comprehensively address the challenges posed by diverse data distributions and non-identically distributed features in federated settings. The proposal is tested through several experiments using four distinct federation strategies: constant-based, accuracy-based, AUC-based, and precision-based weights. The results demonstrate the effectiveness of the approach in achieving the desired learning outcomes while maintaining privacy and confidentiality standards.
\end{abstract}




\end{frontmatter}


\section{Introduction}

Federated learning (FL) is an emerging distributed artificial intelligence framework that enables privacy-preserving machine learning by synthesizing local models instead of sharing actual data \cite{mcmahan.2016}. The general fundamental process can be outlined as follows \cite{salmeron.2023}: the federation process is initiated by a single server or participant who provides an initial model for individual participants to train using their local data. These participants then share the model's weights or gradients with the server (or other participants) for aggregation, typically using the average of the parameters. The federated model is iteratively transmitted back to the participants. This sequence continues until a predefined termination criteria are met. As a result, a federated model trained using the private data contributed by all participants is produced.

This approach is more relevant when addressing sensitive data, particularly in domains like healthcare or finance. Moreover, the need for federated learning with non-independent and identically distributed (non-IID) data has arisen and grown strongly in several industries in recent years. Companies and institutions owning only small and fragmented data have been forming consortia with compensating data partners to collaboratively train machine learning models, but they may have different features space, or these features may have different distributions. In this research, the authors present a new federated learning framework with fuzzy cognitive maps (FCM) that is able to address the challenges posed by diverse data distributions and non-identically distributed features in federated settings.  

In the context of Fuzzy Cognitive Maps (FCMs), federated learning (FL) is employed to address several intrinsic challenges associated with these models. FL offers an effective approach to managing these challenges, enhancing the performance and applicability of FCMs. One necessary issue in the FCMs is the decentralised nature of data sources and the need to preserve data privacy and security while enabling collaborative model deve\-lop\-ment. FCM models frequently rely on data distributed across multiple locations or organisations. Each participant may have unique datasets with varying characteristics, such as different samples and feature spaces. Moreover, datasets may contain sensitive information that participants are hesitant to share due to privacy and security concerns. Building accurate and generalised FCM models requires access to diverse datasets representing different perspectives and contexts.

Therefore, the main contributions of this paper are three-fold:
\begin{itemize}
    \item An approach to distributed learning addressing both horizontal and vertical datasets; hence, it is called square fe\-de\-ra\-ted learning. The introduction of a particle swarm optimisation-based fuzzy cognitive map learning method is presented, specifically designed for datasets containing both shared and distinct, or non-IID, features. To the best of our knowledge, such an approach, where participants have different samples and feature spaces, has not been addressed before. Therefore, this paper focuses on FCMs and the issues resolved with this approach rather than on the accuracy of the training.
    \item A privacy-preserving machine learning strategy for FCMs is proposed. In FCMs, concepts, variables, or features are modelled as nodes, relationships between them as edges, and the weights represent the influence of these relationships. A training framework for collaborative FCM training prioritising data confidentiality was developed. This approach enables multiple participants to jointly train an FCM model using their individual data while ensuring strict compliance with data privacy regulations.
    \item The proposal is tested with four federation strategies: constant-based, accuracy-based, AUC-based, and precision-based weights.
\end{itemize}

This proposal is validated using well-known open datasets. The experimental outcomes confirm the effectiveness of the proposal, demonstrating its successful implementation and its potential to facilitate both horizontal and vertical FL using FCMs. 

The rest of this paper is organised as follows: Section \ref{related.work} delves into the examination of related works. Section \ref{sec:fundamentals} presents the necessary information about fuzzy cognitive maps and federated learning. The methodological proposal is presented in Section \ref{methodological.context}. Detailed insights into the experimental methodology and outcomes are outlined in Section \ref{experiments}. Ultimately, the conclusions can be found in Section \ref{conclusions}.

\section{Related work}\label{related.work}

According to Ma et al. \cite{Ma.2022}, the existing methods to solve the non-IID problem of FL can be enclosed in four areas: data-based, model-based, algorithm-based and framework-based solutions. 

In the data-based space, Zhao \cite{zhao.2018} proposes a previous model trained with shared data in a central server in order to share a subset of shared datasets to each client, while Itahara \cite{itahara.2021} shares unlabeled data to protect privacy. Collins et al. \cite{collins.2021}, train personalised low-dimensional classifiers for each client to learn shared data representations. Other authors propose data augmentation to weaken the effect of statistical heterogeneity \cite{jeong.2023, shin.2020}. 

To alleviate the problems data heterogeneity produces on the model, several authors have introduced changes in the model architecture \cite{sananra.2021}, the aggregation update method \cite{yang.2021}, or the optimisation \cite{leroy.2019, li.2020.heterogeneous}.

With respect to algorithm-based solutions, Jiang \cite{jiang.2023} proposes a model-agnostic meta-learning setting using federated average as the meta-learning algorithm for training and fine-tuning to personalize the model, while Li \cite{li.2022} introduces a meta-learning method for spatio-temporal problems. Smith et al. \cite{smith.2017}, propose a new system-aware optimisation method called MOCHA that learns different tasks and captures their internal relationships in a multi-task setting. 

Finally, some authors have used that the impact of
data heterogeneity on the model can be reduced by client similarity clustering. Ghosh \cite{ghosh.2019} proposes a client-side clustering method that finds an independent local optimal solution for each client, sends it to the server and the server clusters the clients according to that solution. Briggs \cite{briggs.2020} introduces a clustering step after several iterations to hierarchically cluster client based on the similarity between the client’s local update and the global federated model. 

To the best of our knowledge, this work is the first one to exploit the characteristics of fuzzy cognitive maps for solving the heterogeneity problem in federated learning.

\section{Fundamentals}\label{sec:fundamentals}

The integration of fuzzy cognitive maps and federated learning constitutes a synergistic approach that has the potential to significantly advance collaborative machine learning paradigms. FCMs are renowned for their capability to model complex relationships and dynamics within systems, offering a versatile toolset for capturing complex interrelations among variables. Meanwhile, federated learning emerges as an innovative approach to decentralised machine learning, facilitating cooperation among distributed participants while preserving data privacy and security.

In the following sections, the fundamentals of both FCMs and FL will be presented, and an examination will be conducted on how FCMs' ability to encapsulate domain knowledge and capture dynamic behaviours complements the core principles of the FL framework.

\subsection{Fuzzy Cognitive Maps}
Fuzzy cognitive maps are a special case of cognitive maps \cite{axelrod.1976} where the relationships between nodes are modelled with fuzzy cause-effect relations instead of the crisp cause-effect relations proposed by the original cognitive map \cite{salmeron.2006}. The FCM are neuro-fuzzy dynamical systems built from expert knowledge and/or historical raw data. From a computational point of view, FCM models are represented as adjacency matrices, with the weights between the nodes as the elements.

FCMs are composed of nodes that model concepts or variables. Edges represent the relationships between these nodes. The weights represent the influence of these relations \cite{kosko.1986, salmeron.2017}. In other words, the value of a FCM's fuzzy weight $\varpi_{ij}$ describes the influence of node $c_i$ over the node $c_j$. The fuzzy weights between edges are normalised within a range $\mathcal{R}$. If the range is unipolar, then $\mathcal{R}=[0,+1]$ and if it is bipolar, then $\mathcal{R}=[-1,+1]$. The zero value shows that there is no relation between the nodes \cite{lopez.2014, napoles.2021}. 

The calculation of weights is typically performed through the defuzzification of linguistic variables assigned by experts or via machine learning algorithms. In any case, these weights have a semantic significance, and thus, it would not make sense for these weights to be random values as is done in the reservoir computing field.

Moreover, FCMs are intrinsically interpretable methods \cite{rahimi.2022}, unlike other techniques that require the application of explainable artificial intelligence (XAI) approaches for their understanding \cite{guerrero.2020}. Given that FCMs are intrinsically interpretable due to the semantic significance of both the nodes and the weights, it is crucial that the weights and the states of the nodes are not assigned randomly. Random assignment would undermine the semantic integrity and interpretability that are fundamental to FCMs.

The values of the states are computed using activation functions, typically unipolar sigmoid and hyperbolic tangent. These activation functions introduce nonlinearity and saturate the values within the intervals $[0, +1]$ or $[-1, +1]$. Using activation functions like ReLU does not make sense because such functions are linear in the interval $[0, +\infty]$, which would turn an FCM into a purely linear system. Additionally, since ReLU is not bounded within any interval, the system's dynamics could lead to node states with extreme and uncontrolled values.

\subsubsection{Augmented FCMs}
\label{afcm}
FCMs can be built by aggregating multiple FCMs. Several methods have been proposed in literature \cite{salmeron.2009, salmeron.2010, schneider.1998} for FCMs construction. This paper chose to adopt the Augmented FCM approach to build the federated FCM.

The aggregation of each FCM's adjacency matrices enables the creation of an augmented adjacency matrix. The aggregation process is contingent on the presence of shared nodes. The augmented adjacency matrix's ($\odot_{i=1}^N\mathcal{W}_{i}$) states (denoted as $\varpi_{jk}^{FCM}$) are computed by summing the adjacency matrices of each FCM model when common nodes exist within them.

When adjacency matrices do not share common nodes, the summation technique is denoted as the direct sum of matrices, represented as $\odot_{i=1}^{N}\mathcal{W}{i}$. For a any number of FCMs without shared nodes, and even having different numbers of nodes the resultant augmented adjacency matrix is as follows

\begin{equation}
\displaystyle\bigodot_{i=1}^{N}\mathcal{W}_i =
		\left[
			\begin{array}{cccc}
				 0 & \cdots & 0 &  \mathcal{W}^A_{1\times k}  \\[0.3cm]
                 0 & \cdots &   \mathcal{W}^B_{2\times k-1} & 0  \\[0.3cm]
                 \vdots & \ddots & \vdots &  \vdots  \\[0.3cm]
				\mathcal{W}^Z_{z\times 1}  &  \cdots & 0 & 0
			\end{array}
		\right]
\end{equation}

\noindent where $N$ is the number of adjacency matrices to merge, zeros are actually zero matrices, and the dimension of $\displaystyle\odot_{i=1}^{N}\mathcal{W}_{i}$ is $[\,\cdot\,]_{(z+k)\times(z+k)}$. Note that $z+k$ represents the total count of distinct nodes across all adjacency matrices. 

In the case of common nodes, they would be computed as the average (or any other aggregation method) of the nodes' states in each adjacency matrix $\mathcal{W}_{i}$. The computation is shown as Eq. \ref{eq:aug2} where $\oplus(\cdot)$ is the selected aggregation method (usually addition).

\begin{figure*}
\begin{equation}
\bigodot_{i=1}^{M}\mathcal{W}_i =
		\left[
			\begin{array}{ccc}
				 \displaystyle\frac{1}{M}\cdot\displaystyle\bigoplus_{m=1}^M\left\{
	\begin{array}{ll}
\mathcal{W}^m_{1\times 1}  & if\;  \exists {W}^m_{1\times 1} \\[0.2cm]
                0 &  otherwise
	\end{array}\right.  & \cdots &  \displaystyle\frac{1}{M}\cdot\displaystyle\bigoplus_{m=1}^M\left\{
	\begin{array}{ll}
	\mathcal{W}^m_{1\times k}  & if\;  \exists {W}^m_{1\times k}\\[0.3cm]
                0 &  otherwise
	\end{array}\right.  \\[0.3cm]
                 \vdots & \ddots &  \vdots  \\[0.3cm]
					\displaystyle\frac{1}{M}\cdot\displaystyle\bigoplus_{m=1}^M\left\{
	\begin{array}{ll}
\mathcal{W}^m_{z\times 1}  & if\;  \exists {W}^m_{z\times 1}\\[0.2cm]
                0 &  otherwise
	\end{array}\right.  &  \cdots & 	\displaystyle\frac{1}{M}\cdot\displaystyle\bigoplus_{m=1}^M\left\{
	\begin{array}{ll}
\mathcal{W}^m_{z\times k}  & if\;  \exists {W}^m_{z\times k}\\[0.2cm]
                0 &  otherwise
	\end{array}\right. 
			\end{array}
		\right]
\label{eq:aug2}
\end{equation}
\end{figure*}

Recently, other authors have developed different aggregations methods for FCMs, for instance the CDFCM (Causal Discovery from FCM) model \cite{teng.2024}.

%
%
%
%
%
%

\subsubsection{FCM learning}
Fuzzy cognitive maps learning efforts typically focus on the automatic construction of the adjacency matrix using either historical raw data or expert knowledge \cite{salmeron.2019}. FCM learning approaches can be categorised into four types: Hebbian, population-based, hybrid (combining Hebbian and population-based learning algorithms \cite{napoles.2021, napoles.2023, salmeron.2017b}) and emerging approaches. Table \ref{tab:learning} shows the main FCM learning approaches.

Hebbian-based FCM learning approaches aim to update adjacency matrices in order to guide the FCM model towards achieving a steady state or converging into an acceptable region for the target system. However, these approaches have not been successful in the case of FCM extensions like FGCMs \cite{salmeron.2010, froelich.2014, salmeron.2019b, salmeron.2021}.

\begin{table}[t]  
\scriptsize
\centering 
\caption{FCM Learning Algorithms} 
\label{tab:learning} 
\setlength{\tabcolsep}{1pt} 
	\begin{tabular}{p{1.5cm}p{6.0cm}l} 
		\hline
		Category & Learning approach & Author(s) \\ 
		\hline\hline
		Hebbian & Differential Hebbian Learning (DHL/DDNHL) &  \cite{dickerson1994virtual, kosko1996fuzzy, salmeron.2019b, stach2008data} \\
	    & Nonlinear Hebbian Learning (NHL) & \cite{papageorgiou2003fuzzy, salmeron.2019b} \\
     	& Balance Differential Algorithm (BDA) & \cite{huerga2002balanced, salmeron.2019b} \\
     	& Petri Nets & \cite{konar2005reasoning} \\ 
		& Active Hebbian Learning (AHL) & \cite{papageorgiou2004active} \\ \hline
		Population & Evolutionary Strategies (ES) & \cite{koulouriotis2001learning} \\
  			& Genetic Algorithm (GA/RCGA/SOGA) &  \cite{mateou2005multi, stach2005genetic, poczketa2015learning} \\
			& Particle Swarm Optimisation (PSO) & \cite{parsopoulos2003first, salmeron.2017} \\
			& Memetic Particle Swarm optimisation (MPSO) & \cite{petalas2005fuzzy, salmeron.2017b} \\
			& Simulated Annealing (SA/CSA) & \cite{ghazanfari2007comparing, alizadeh2009learning} \\
			& Tabu Search (TS) & \cite{alizadeh2007learning} \\
			& Game-based learning & \cite{luo2009game} \\
			& Differential Evolution (DE) & \cite{juszczuk2009learning} \\
			& Immune Algorithm & \cite{lin2009immune} \\
			& Big-Bang Big-Crunch (BB-BC) & \cite{yesil2010big} \\
			& Self-Organized Migration Algorithm (SOMA) & \cite{vavsvcak2010approaches} \\
			& Ant Colony optimisation (ACO) & \cite{ding2011first} \\
			& Extended Great Deluge Algorithm (EGDA) & \cite{baykasoglu2011training} \\
			& Artificial Bee Colony (ABC) & \cite{yesil2013fuzzy} \\
			& Cultural Algorithm (CA) & \cite{ahmadi2014first} \\
			& Imperialist Competitive Algorithm (ICLA) & \cite{ahmadi2015learning} \\
			& Modified Asexual Reproduction & \cite{salmeron.2019}\\
			& Multiobjective Evolutionary Algorithm (MOEA-FCM) & \cite{chi2016learning} \\ \hline
		Hybrid & NHL-DE & \cite{papageorgiou2005new} \\
			& NHL-RCGA & \cite{zhu2008integrated} \\
			& NHL-EGDA & \cite{ren2012learning} \\ \hline
		Emerging & Pseudoinverse learning & \cite{Vanhoenshoven.2020} \\ 
        & Federated Learning & \cite{salmeron.2023, salmeron.2024} \\
        & Quantum-based learning & \cite{rahimi.2022} \\\hline
	\end{tabular} 
\end{table}

Population-based methods (e.g.: Particle Swarm Optimisation, PSO) eliminate the need for human intervention as they build adjacency matrices from historical raw data that effectively capture the sequence of input state vectors (dataset ins\-tan\-ces). The objective of evolutionary learning in FCMs is to generate an optimal adjacency matrix that accurately models the behaviour of a system.

%

PSO is a bio-inspired optimisation algorithm that utilises a population-based approach with stochastic elements. PSO algorithm creates a swarm of particles that move within an $n$-dimensional search space, encompassing all possible candidate solutions. The training of the adjacency matrices of the FCM involves considering the position of the $k$-th particle (a candidate solution or adjacency matrix), denoted as $\varpi_k=[\varpi_{k_1},\ldots,\varpi_{k_j}]$, along with its velocity, $v_k=[v_{k_1},\ldots,v_{k_j}]$. Note that each particle represents a potential solution or candidate for the FCM, and its position $\varpi_k$ corresponds to the adjacency matrix of the $k$-th FCM candidate \cite{salmeron.2017}. 

At each time step, the velocity and position of each particle are updated. The computations for particle position and velocity are determined by Eqs. \ref{pso_a} and \ref{pso_b} respectively. In Eq. \ref{pso_a}, $U(0,\phi_i)$ represents a vector of random numbers generated from a uniform distribution within the range of $[0,\phi_i]$. This vector is generated for each iteration and each particle. Additionally, $\dot{\varpi}_k$ corresponds to the best position of particle $k$ achieved in previous iterations, $\ddot{\varpi}_k$ represents the best position attained by the entire population in all previous iterations, and $\otimes$ denotes component-wise multiplication.
\begin{equation}
\label{pso_a}
\varpi_k(t+1) = \varpi_k(t) + v_k(t)
\end{equation}
\begin{equation}
\label{pso_b}
\begin{split}
v_k(t+1) = v_k(t) &+ U(0,\phi_1)\otimes \Big(\dot{\varpi}_k -\varpi_k(t)\Big) \\
& + U(0,\phi_2)\otimes \Big(\ddot{\varpi}_k -\varpi_k(t)\Big) 
\end{split}
\end{equation}

%
%


\subsection{Federated Learning}
Federated learning is a distributed way of training an artificial intelligence model using the data of se\-ve\-ral participants in a secure way. It was proposed by McMahan et al. \cite{mcmahan.2016} and further developed in Konecny et al. \cite{konecn.2016} and McMahan and Ramage \cite{mcmahan_ramage.2017} for deep learning models, but has been extended to fe\-de\-ra\-ted linear regression \cite{Karr2009PrivacyPreservingAO, Gascon.2017}, federated logistic regression \cite{Hardy2017PrivateFL}, federated random forest \cite{Liu2019FederatedF}, federated XGBoost \cite{Cheng2019SecureBoostAL, Fang2020AHF, XieFedXGB}, federated support vector machine \cite{Gu2020FederatedDS, Yu2006PrivacyPreservingSC}, and federated FCMs \cite{salmeron.2023, salmeron.2024, salmeron.2020}. 

\begin{figure}[!ht]
\centerline{\includegraphics[width=7cm]{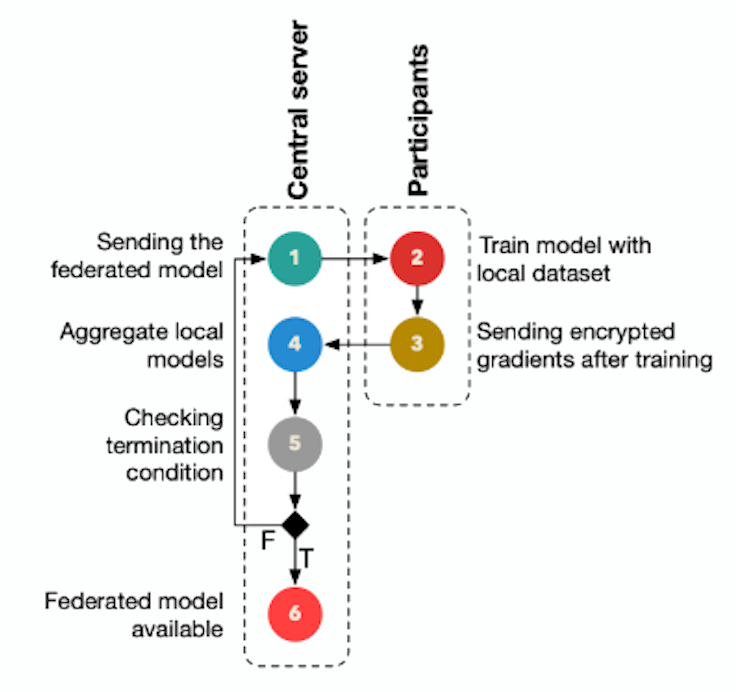}}
\caption{A federated learning process. Adapted from \cite{salmeron.2024}.} \label{FL}
\end{figure}

The process of training a federated model (Figure \ref{FL}) unfolds as follows:
\begin{enumerate}
\item The central server (or a designated node in a P2P architecture) delivers the model, which in the initial iteration is an empty model, to every agent.
\item The participants train the model with their own private data.
\item Each participant sends the parameters of the model or its gradients to the central server in a private way, usually encrypted.
\item The central server builds a federated model by aggregating the parameters of the individual models. Different aggregation methods will optimise the federated model in different ways (see Subsection \ref{aggregation.methods}). 
\item The central server checks if the termination condition is met. If it is, the federated model is finished; otherwise, the process returns to step 1.
\end{enumerate}

The aim of the federated model is to minimise the total loss across the data of all participants:
\begin{equation}
     \mathcal{L}^* = \sum_{i=1}^n \kappa_i\cdot %
     \mathcal{L}(\mathcal{D}_i,\Psi) 
\end{equation}

\noindent where $\Psi$ are the model parameters, $\mathcal{D}_i$ is the dataset of the participant $i$, $\mathcal{L}^*$ is the loss function of the federated model, $\mathcal{L}_i(\cdot)$ is the loss function for each participant's model in the federation, and $\kappa_i$ represents the weight of each participant, which can be determined by criteria such as dataset size, accuracy, and so on. 


This federated process entails training a model across various participants without the need to share private data. However, the transmission of data introduces potential risks such as model poisoning or attacks aimed at reconstructing the model or training data. Consequently, there has been significant progress in employing privacy-preserving techniques like di\-ffe\-ren\-tial privacy, homomorphic encryption or chaotic maps encryption  \cite{pmlr-v108-bagdasaryan20a, Wang19, abadiDP16, Acar17, arevalo23}.

\subsubsection{Federated Learning Categories}

\begin{figure*}[!ht]
\centerline{\includegraphics[width=18cm]{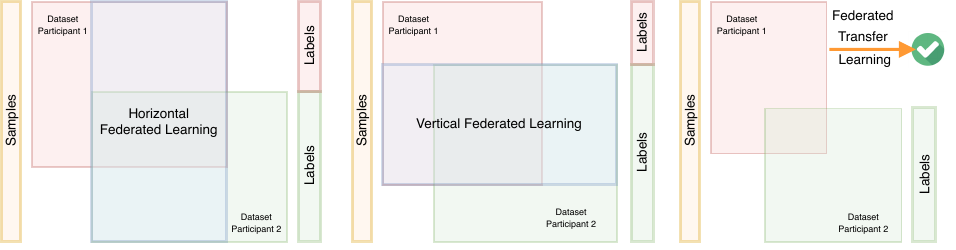}}
\caption{Federated Learning categories} 
\label{FL.types}
\end{figure*}

There are three kinds of federated learning projects depending on the nature of the data \cite{li.2020,yang.2019}: horizontal fede\-ra\-ted learning, vertical federated learning and federated transfer learning (Figure \ref{FL.types}).

Horizontal federated learning is the subset of federated learning where the features space is shared by all participants, but the samples space is different. This approach constitutes the initial proposal for federated learning, yet it continues to pose certain challenges. For instance, a novel strategy known as hierarchical heterogeneous horizontal federated learning encounters limitations due to the scarcity of labeled entities in the context of horizontal federated learning \cite{li.2020}. In this research, the scarcity of labeled instances is alleviated by ite\-ra\-ti\-vely adjusting the heterogeneous domain, with each participant serving as the target domain in successive iterations.

In vertical federated learning, or heterogeneous federated learning, each participant in a common sample space possesses a distinct feature space. Unlike in horizontal federated learning, consolidating the entire dataset into one participant to train a global model is not feasible in vertical FL \cite{, salmeron.2024, cheng.2019,lee.2020}. In the final case, when the data does not share a sample space or a feature space, the federated transfer learning approach proposed by \cite{Liu.2020} applies to common parties with small data intersection to build a model using transfer learning in a distributed way. 

This proposal can be readily adjusted to diverse secure machine learning endeavours with only minor modifications to the existing model, while still delivering comparable accuracy as non-privacy-preserving transfer learning.

\subsubsection{Non-IID data for Federated Learning}

Since each participant may have obtained their data from different sources, the distributions of the data could be quite dissimilar from each other. This situation is known as non-IID \cite{mcmahan.2017}. This fact is another challenge for the training of a federated model. Zhe et al. \cite{zhu.2021} propose the following categories of non-IID data:
\begin{itemize}
    \item \underline{Attribute skew}, that includes these subcategories: 
    \begin{itemize}
        \item \textit{Non-overlapping attribute skew}: The data attributes for the clients are mutually exclusive.
        \item \textit{Partial overlapping attribute skew}: Portions of the data attributes can be shared.
        \item \textit{Full overlapping attribute skew}: Data attributes are the same in all participants but the distributions are different.
    \end{itemize}
    \item \underline{Label skew}, with the subcategories: 
    \begin{itemize}
        \item \textit{Label distribution skew}: Label distributions on the clients are different.         \item \textit{Label preference skew}: The label distribution is different on the client data, although the distribution of the attributes is the same.         
        \item \textit{Temporal skew}: There is distribution skewness in temporal data, which encompasses spatio-temporal data as well as time series data.
    \end{itemize}
    \item \underline{Attribute and label skew}. Different clients have data with different labels and different attributes.
    \item \underline{Quantity skew}. The number of training data varies across different clients.
\end{itemize}

\section{Methodological proposal}
\label{methodological.context}
In this section, the square federated learning approach (Section \ref{fcm_fl}) and the aggregation methods (Section \ref{aggregation.methods}) are detailed. Both sections are the core of our methodological proposal.

\subsection{Square Federated Learning for FCMs} 
\label{fcm_fl}

Our square federated learning proposal combines horizontal and vertical federated learning, enabling model training when both the sample space and feature space vary across participants. This proposal is also suitable for non-IID data, especially partial overlapping attribute skew. This is achieved through the characteristics of the FCMs, particularly the aggregation with augmented adjacency matrices (Eq. \ref{eq:aug2}), because a federated augmented FCM can be built even in the case when the participants have non-iid data. This enables training with minimal to no overlapping in the attribute space by creating a new FCM model.

The square federated learning methodology for FCMs is as follows:
\begin{enumerate}
    \item The central server triggers the process by sending the model to the participants, who own the data to train the final model. This step and the aggregation could actually be done by one of the participants or even all of them, making the central server not even necessary.
    \item Each participant trains its own initial local FCM with their own dataset. This research have used a PSO-based FCM learning approach, and the convergence condition as the difference between two consecutive vector states being less than a tolerance value of $1\times 10^{-5}$, but this proposal is agnostic to these parameters. 
    \item Each participant sends the parameters of its trained model —in this case, the locally trained adjacency matrices— to the central server, while the local FCM is stored on the participant devices for evaluation. 
    \item The central server aggregates the parameters of the local models in its device using the appro\-priate weight. Subsection \ref{aggregation.methods} shows a detailed description of the different aggregation methods considered in this research. The output of this process are the parameters of the federated model. 
    \item The participants receive the parameters from the central server and build the next iteration of their local model, in this case they receive the adjacency matrices and build an FCM model.
    \item Every participant retrains the new FCM in their local data and evaluate the performance of their local model in a test set. They also send the parameters back to the central server to be aggregated once again. 
    \item The central server checks whether the termination condition is met. If the condition is not fulfilled, the process goes back to stage 4, if it is satisfied, then the Fede\-rated learning process ends resulting in a federated FCM. 
\end{enumerate}

The difference between square federated learning and the conventional FL framework lies in the versatility of fuzzy cognitive maps. It is a scientific challenge to extend square FL to other models, and this requires further research. FCMs are particularly suitable for federated learning due to their flexible aggregation methods (Subsection \ref{aggregation.methods}). Moreover, FCMs are able to aggregate different models into a augmented FCM with varying concepts and relationships, as in the case of non-iid data in federated learning.

\subsection{Aggregation methods for Federated Learning}
\label{aggregation.methods}

Federated learning aggregation approach is an important ele\-ment of the process. The original definition contemplated the arithmetic mean of the parameters of the model to obtain the federated model \cite{mcmahan.2016}. 

This research proposes four different aggregations and compare them in our experimental section (see Section \ref{experiments}). Given the parameters of the local models at iteration $j$, denoted as $\Phi_{j}=[\Phi_{j1}, \Phi_{j2}, \cdots, \Phi_{jn}]$, where $n$ represents the number of participants, $\Phi_{ji}$ represents the parameters of the local model for participant $i$ at iteration $j$, and $\Phi_j'$ represents the parameters (adjacency matrices in FCM models) of the federated model, the functions of the parameters that will be discussed are as follows: 
\begin{itemize}
    \item Average of the parameters (weights or gradients):
    \begin{equation}\label{average}
    \Phi_j' = \frac{1}{n}\sum_{i=1}^n \Phi_{ji}
    \end{equation}
    where every participant contributes the same to the global model.
    \item Accuracy-based weights: Weighted average using the normalised accuracy of the model computed in a test set of each participants,
    \begin{equation}\label{accuracy}
    \Phi_j' = \sum_{i=1}^n \frac{\textrm{acc}_{ji}}{\sum_{k = 1}^n \textrm{acc}_{jk}}\cdot\Phi_{ji}
    \end{equation}
    \noindent where $\textrm{acc}_{ji}$ is the accuracy of the local model of par\-ti\-ci\-pant $i$ in a test set at iteration $j.$ Using this averaging strategy, the individual models contribute to the global model inversely proportional to their performance metric. This approach aims to give more weight to the less accurate models to improve their performance on their respective datasets. 
    \item AUC-based weights: Weighted average using the normalised Area Under the Curve (AUC) of the model in a test set of each participants,
    \begin{equation}\label{AUC}
    \Phi_j' = \sum_{i=1}^n \frac{\textrm{AUC}_{ji}}{\sum_{k = 1}^n \textrm{AUC}_{jk}}\cdot\Phi_{ji}
    \end{equation}
    \noindent where $\textrm{AUC}_{ji}$ is the AUC of the local model of participant $i$ in a test set at iteration $j$. Here, the individual models contribute to the global model inversely proportional to their performance metric, giving more weight to the mo\-dels with lower AUC to improve their metrics on their res\-pec\-tive datasets. 
    \item Precision-based weights: Weighted average using the normalised precision of the model in a test set of each participants,
    \begin{equation}\label{precision}
    \Phi_j' = \sum_{i=1}^n \frac{\textrm{prec}_{ji}}{\sum_{k = 1}^n \textrm{prec}_{jk}}\cdot\Phi_{ji}
    \end{equation}
    \noindent where $\textrm{prec}_{ji}$ is the precision of the local model of participant $i$ in a test set at iteration $j.$ With this federation the individual models add to the global model inversely to their performance metric, trying to give more weight to the less precise models in order to improve their metric in their datasets. 
\end{itemize}

\section{Experimental approach} 
\label{experiments}
The scenario for the experiments is as follows. It is assumed that there are several participants with private data who wish to train an FCM classification model. However, the amount of available data for each participant is not enough for training a robust model. The selected initial model is an FCM with sigmoid activation function and trained with a PSO using 50 iterations and a swarm size of 10 particles. 

For the experimental approach, it is assumed that there are five different participants, each with the same amount of data, obtained from an evenly split dataset for horizontal federated learning. In the vertical federated learning experiments, each participant's dataset will have three features randomly removed. 

Each of the experiments will use one of the aggregations discussed in Subsection \ref{aggregation.methods}: constant-based, accuracy-based, AUC-based, and precision-based weights. 

The results are presented in tables with the average accuracy and F1 metrics for all participants. The columns are as follows: the name of the dataset, the mean accuracy for the initial FCM model (pre-FL), the mean accuracy after the FL process, the mean F1 score of the initial model, and the mean F1 score after the federation. These metrics will be computed using a test dataset in each participant.

The data will be obtained from the classification datasets from Penn machine learning Benchmarks \cite{romano2021pmlb, Olson2017PMLB}. The interested reader can access all of these datasets by installing the package \code{pmlb} in Python or R languages. All datasets presented in this research consist of tabular data and represent binary classification problems.


\subsection{Baseline: Centralised models}

A centralised FCM and a centralised neural network with 6 hidden layers and dropout are trained with all data stored on a single node to serve as a baseline for comparison. Table \ref{blind.Centralised} displays the accuracy and F1-score metrics for the centralised models, although it should be mentioned that the aim of this research is not to improve the performance of these models. Our setting involves the distributed environment in which several participants would not be able to train them due to the sharing of the data and the different distributions of features. 

\begin{table}[ht]
\footnotesize
\centering
\caption{Centralised baseline}\label{blind.Centralised}
\begin{tabular}{|l|| c | c || c | c |}
\hline
 & \textbf{FCM} & \textbf{FCM} & \textbf{NN} & \textbf{NN}\\
\textbf{Dataset} & \textbf{Accuracy} & \textbf{F1 score} & \textbf{Accuracy} & \textbf{F1 score}\\
\hline \hline
adult & 0.8632 & 0.87 & 0.8527 & 0.8984 \\
australian & 0.9420 & 0.9403 & 0.8478 & 0.8969 \\
breast cancer & 0.7931 & 0.6250 & 0.7241 & 0.6741 \\
buggyCrx & 0.8333 & 0.8535 & 0.8478 & 0.8096 \\
chess & 0.5646 & 0.6874 & 0.5937 & 0.6017 \\
credit\_a  & 0.6971 & 0.6646 & 0.8478 & 0.6195 \\
credit\_g & 0.8333 & 0.8780 & 0.8000 & 0.8281  \\
crx & 0.9050 & 0.8923 & 0.8478 & 0.8720 \\
german & 0.6998 & 0.8213 & 0.8000 & 0.7889 \\
heart\_c & 0.7733 & 0.7892 & 0.8525 & 0.7723 \\
heart\_h & 0.8009 & 0.8406 & 0.7627 & 0.7735 \\
heart\_statlog & 0.9103 & 0.8945 & 0.9148 & 0.8638 \\
house\_votes\_84 & 0.8722 & 0.8405 & 0.9655 & 0.8892 \\
hungarian & 0.9355 & 0.9091 & 0.8475 & 0.8814 \\
hypothyroid & 0.9537 & 0.9763 & 0.9684 & 0.9666 \\
kr\_vs\_kp & 0.6094 & 0.6383 & 0.5891 & 0.6010 \\
parity5+5 & 0.6790 & 0.6061 & 0.6444 & 0.6010 \\
profb & 0.8518 & 0.7309 & 0.8370 & 0.7457 \\
saheart & 0.9290 & 0.8812 & 0.8989 & 0.8618 \\
threeOf9 & 0.9236 & 0.9173 & 0.8544 & 0.8043 \\
tic\_tac\_toe & 0.7670 & 0.8253 & 0.7604 & 0.7721 \\
\hline
\end{tabular}
\end{table}

\subsection{Experiment 1: Constant weights}

In this experiment, a square federated learning approach with constant weights is employed across twenty-two open classification datasets. As previously mentioned, two metrics are compared: accuracy and F1 score. The mean performance of the initial model across five participants improves after the federation process in all cases in terms of accuracy, while the F1 score increases in all but 2 out of 22 cases (see Table \ref{blind.constant}) 

The increase in accuracy is expected because averaging the models during the federation process generally enhances the overall performance of the models on the participants' datasets. However, it is observed that the federation may decrease the F1 score for some cases, particularly when using an aggregation method that does not rely on any metric involved in the F1 score. 

It is also noteworthy that the setting of the FCMs, with a large number of possible iterations and sigmoid activation function, is able to reach performance results comparable to the centralised case. Nevertheless, the goal of this research is not to enhance the metrics of this model, but to find a federated model even with different features distributions across the participants.


\begin{table}[ht]
\footnotesize
\centering
\caption{Square FL results (constant weights)}\label{blind.constant}
\begin{tabular}{|l|| c| c|| c | c |}
\hline
 & \textbf{Mean acc.} & \textbf{Mean acc.} & \textbf{Mean F1} & \textbf{Mean F1}\\
\textbf{Dataset} & \textbf{pre-FL} & \textbf{post-FL} & \textbf{pre-FL} & \textbf{post-FL}\\
\hline \hline
adult & 0.8111 & \textbf{0.8642} & 0.8823 & \textbf{0.8902} \\
australian & 0.8706 & \textbf{0.9212} & 0.8722 & \textbf{0.9135} \\
breast cancer & 0.9208 & \textbf{0.9356} & 0.5503 & \textbf{0.5934} \\
buggyCrx & 0.8122 & \textbf{0.8464} & 0.7993 & \textbf{0.8034} \\
chess & 0.6973 & \textbf{0.8193} & 0.7141 & \textbf{0.7352} \\
credit\_a & 0.6491 & \textbf{0.7081} & \textbf{0.6362} & 0.5936 \\
credit\_g & 0.7511 & \textbf{0.8756} & 0.8010 & \textbf{0.8420} \\
crx & 0.8143 & \textbf{0.8897} & 0.8566 & \textbf{0.8712} \\
german & 0.7555 & \textbf{0.8325} & 0.8276 & \textbf{0.8571} \\
heart\_c & 0.7319 & \textbf{0.7664} & 0.7575 & \textbf{0.7923} \\
heart\_h & 0.7927 & \textbf{0.8203} & 0.8012 & \textbf{0.8224} \\
heart\_statlog & 0.7996 & \textbf{0.8891} & 0.7441 & \textbf{0.7869} \\
house\_votes\_84 & 0.8467 & \textbf{0.8894} & 0.8080 & \textbf{0.8325} \\
hungarian & 0.6897 & \textbf{0.8551} & 0.6507 & \textbf{0.7714} \\
hypothyroid & 0.9337 & \textbf{0.9647} & 0.9645 & \textbf{0.9763} \\
kr\_vs\_kp & 0.5650 & \textbf{0.6726} & 0.6380 & \textbf{0.6632} \\
parity5+5 & 0.6123 & \textbf{0.6667} & \textbf{0.5939} & 0.5835 \\
profb & 0.8125 & \textbf{0.8409} & 0.7213 & \textbf{0.7455} \\
saheart & 0.9212 & \textbf{0.9215} & 0.8556 & \textbf{0.8615} \\
threeOf9 & 0.8932 & \textbf{0.9134} & 0.9091 & \textbf{0.9248} \\
tic\_tac\_toe & 0.7379 & \textbf{0.7522} & 0.8023 & \textbf{0.8124} \\
\hline
\end{tabular}
\end{table}

\subsection{Experiment 2: Accuracy-based weights}

The results for the second experiment, using an accuracy-based aggregation method, can be found in Table \ref{blind.accuracy}. 

As in the previous case, accuracy improves in every instance, as expected, since the accuracy-based federated process optimises the aggregation of partial models to enhance the accuracy of all models and achieve a higher mean accuracy. However, theoretically, this aggregation does not improve the F1-score. In fact, for 3 out of 22 datasets in the experiments conducted, this metric is worse. This result is to be expected since this federation process would optimize the accuracy metric, which the plain FL system already handles, and does nothing for the optimisation of the F1 score.

In comparison with the centralised cases, several instances of better mean performance for the federated model than the single model case are observed.

\begin{table}[ht]
\footnotesize
\centering
\caption{Square FL results (accuracy-based weights)}\label{blind.accuracy}
\begin{tabular}{|l| c| c| c | c |}
\hline
 & \textbf{Mean acc.} & \textbf{Mean acc.} & \textbf{Mean F1} & \textbf{Mean F1}\\
\textbf{Dataset} & \textbf{pre-FL} & \textbf{post-FL} & \textbf{pre-FL} & \textbf{post-FL}\\
\hline
adult & 0.8205 & \textbf{0.8432} & \textbf{0.8333} & 0.8111 \\
australian & 0.8575 & \textbf{0.9022} & 0.9011 & \textbf{0.9367} \\
breast cancer & 0.9113 & \textbf{0.9309} & 0.5822 & \textbf{0.6189}\\
buggyCrx & 0.7887 & \textbf{0.7902} & 0.7571 & \textbf{0.7809}\\
chess & 0.6004 & \textbf{0.7017} & 0.6819 & \textbf{0.7134} \\
credit\_a & 0.5735 & \textbf{0.7624} & 0.4273 & \textbf{0.6313} \\
credit\_g & 0.6537 & \textbf{0.8067} & 0.7440 & \textbf{0.7881}\\
crx & 0.8245 & \textbf{0.8699} & 0.7738 & \textbf{0.8383} \\
german & 0.7240 & \textbf{0.8097} & 0.8150 & \textbf{0.8560} \\
heart\_c & 0.7391 & \textbf{0.7412} & \textbf{0.7692} & 0.7189 \\
heart\_h & 0.7679 & \textbf{0.7934} & 0.7533 & \textbf{0.8102} \\
heart\_statlog & 0.8118 & \textbf{0.8529} & 0.7478 & \textbf{0.7560} \\
house\_votes\_84 & 0.9692 & \textbf{0.9846} & 0.9500 & \textbf{0.9560} \\
hungarian & 0.7125 & \textbf{0.8476} & \textbf{0.6111} & 0.5991 \\
hypothyroid & 0.9337 & \textbf{0.9647} & 0.9645 & \textbf{0.9763} \\
kr\_vs\_kp & 0.6340 & \textbf{0.7472} & 0.6640 & \textbf{0.7405} \\
parity5+5 & 0.6731 & \textbf{0.7469} & 0.5928 & \textbf{0.6070} \\
profb & 0.8044 & \textbf{0.8367} & 0.7000 & \textbf{0.7267} \\
saheart & 0.9092 & \textbf{0.9256} & 0.8455 & \textbf{0.8609} \\
threeOf9 & 0.8812 & \textbf{0.9046} & 0.8734 & \textbf{0.9048} \\
tic\_tac\_toe & 0.7454 & \textbf{0.7778} & 0.7400 & \textbf{0.7986}  \\

\hline
\end{tabular}
\end{table}

\subsection{Experiment 3: AUC-based weights}

The experiment involving AUC-based federated learning demonstrates an improvement in mean accuracy across all cases following the federation process, aligning with the observations made after experiment 1. Also once again the federation is able to improve the accuracy of the Centralised model. However, the weights in this experiment are not optimised to enhance the mean F1-score. 

Table \ref{blind.auc} shows that the number of datasets for which the mean F1-score is not improved after the federation process is 9 out of 22.

\begin{table}[ht]
\footnotesize
\centering
\caption{Square FL results (AUC-based weights)}\label{blind.auc}
\begin{tabular}{|l| c| c| c | c |}
\hline
 & \textbf{Mean acc.} & \textbf{Mean acc.} & \textbf{Mean F1} & \textbf{Mean F1}\\
\textbf{Dataset} & \textbf{pre-FL} & \textbf{post-FL} & \textbf{pre-FL} & \textbf{post-FL}\\
\hline
adult &  0.8125 & \textbf{0.8562} & 0.8224 & \textbf{0.8681} \\
australian & 0.8635 & \textbf{0.9103} & 0.8767 & \textbf{0.9145} \\
breast cancer & 0.9208 & \textbf{0.9425} & 0.6108 & \textbf{0.6378} \\
buggyCrx & 0.7825 & \textbf{0.8213} & 0.8023 & \textbf{0.8178} \\
chess & 0.6262 & \textbf{0.7297} & 0.6769 & \textbf{0.7169} \\
credit\_a & 0.6581 & \textbf{0.6766} & \textbf{0.6277} & 0.5842 \\
credit\_g & 0.8035 & \textbf{0.9097} & 0.8600 & \textbf{0.9124} \\
crx & 0.7885 & \textbf{0.8589} & 0.6671 & \textbf{0.7651} \\
german & 0.7940 & \textbf{0.8503} & \textbf{0.8646} & 0.8642 \\
heart\_c & 0.7047 & \textbf{0.7809} & \textbf{0.8106} & 0.7754 \\
heart\_h & 0.7056 & \textbf{0.8174} & \textbf{0.8054} & 0.7823 \\
heart\_statlog & 0.8183 & \textbf{0.8700} & 0.7442 & \textbf{0.7662} \\
house\_votes\_84 & 0.8737 & \textbf{0.9185} & 0.8441 & \textbf{0.8558} \\
hungarian &  0.8167 & \textbf{0.9373} & 0.7592 & \textbf{0.8758}  \\
hypothyroid & 0.9515 & \textbf{0.9525} & \textbf{0.9750} & 0.9715 \\
kr\_vs\_kp &  0.6144 & \textbf{0.7146} & 0.6438 & \textbf{0.6837} \\
parity5+5 &  0.6616 & \textbf{0.6885} & \textbf{0.5843} & 0.5400  \\
profb &   0.8388 & \textbf{0.8521} & 0.6908 & \textbf{0.7154} \\
saheart &  0.8801 & \textbf{0.8947} & \textbf{0.9015} & 0.8734   \\
threeOf9 &  0.9284 & \textbf{0.9432} & \textbf{0.9027} & 0.8934  \\
tic\_tac\_toe & 0.7337 & \textbf{0.7485} & \textbf{0.8302} & 0.8092\\
\hline
\end{tabular}
\end{table}

\subsection{Experiment 4: Precision-based weights}

Finally, the last experiment uses an aggregation method based in the precision metric of each participant's dataset. 

As in all previous cases, the mean accuracy is improved by the federation process. Regarding the F1 score, since this metric is related to the optimised metric in this experiment, precision. The number of datasets for which the F1 score increases after the federated learning approach is all but 4 out of 22 — a better result than for the experiments with AUC-based weights. Ho\-we\-ver, it is not higher that the improvement seen in the first and second experiments with constant and accuracy-based weights.

Again, the averaging using the precision metric does not prevent the mean accuracy to improve the centralised-model accuracy in several cases. 

\begin{table}[ht]
\footnotesize
\centering
\caption{Square FL results (precision-based weights)}\label{blind.precision}
\begin{tabular}{|l| c| c| c | c |}
\hline
 & \textbf{Mean acc.} & \textbf{Mean acc.} & \textbf{Mean F1} & \textbf{Mean F1}\\
\textbf{Dataset} & \textbf{pre-FL} & \textbf{post-FL} & \textbf{pre-FL} & \textbf{post-FL}\\
\hline
adult &  0.7944 & \textbf{0.8667} & 0.8654 & \textbf{0.8784} \\
australian & 0.8352 & \textbf{0.8805} & 0.7614 & \textbf{0.8566} \\
breast cancer & 0.8547 & \textbf{0.9134} & 0.5809 & \textbf{0.5933} \\
buggyCrx & 0.7735 & \textbf{0.8183} & \textbf{0.8239} & 0.8062 \\
chess & 0.6835 & \textbf{0.7614} & 0.7238 & \textbf{0.7299} \\
credit\_a & 0.4788 & \textbf{0.6193} & 0.4480 & \textbf{0.6213} \\
credit\_g &  0.6811 & \textbf{0.7524} & 0.7645 & \textbf{0.8135} \\
crx & 0.7974 & \textbf{0.8450} & 0.6567 & \textbf{0.7117} \\
german & 0.7562 & \textbf{0.8269} & 0.8256 & \textbf{0.8650} \\
heart\_c & 0.7633 & \textbf{0.7839} & 0.7995 & \textbf{0.8122} \\
heart\_h & 0.7611 & \textbf{0.7988} & 0.8112 & \textbf{0.8273} \\
heart\_statlog & 0.8179 & \textbf{0.8787} & 0.7681 & \textbf{0.7925} \\
house\_votes\_84 & 0.8982 & \textbf{0.9559} & 0.8660 & \textbf{0.8731} \\
hungarian &  0.8170 & \textbf{0.9314} & 0.7109 & \textbf{0.7316} \\
hypothyroid & 0.7458 & \textbf{0.8805} & 0.8305 & \textbf{0.9235} \\
kr\_vs\_kp &  0.6583 & \textbf{0.7165} & \textbf{0.7373} & 0.7126 \\
parity5+5 &  0.6178 & \textbf{0.6932} & \textbf{0.5717} & 0.5534 \\
profb & 0.8407 & \textbf{0.8648} & 0.7333 & \textbf{0.7611} \\
saheart &  0.9091 & \textbf{0.9268} & 0.8854 & \textbf{0.9012} \\
threeOf9 &  0.8732 & \textbf{0.8954} & 0.9012 & \textbf{0.8634} \\
tic\_tac\_toe & 0.7890 & \textbf{0.8345} & \textbf{0.8534} & 0.8255 \\
\hline
\end{tabular}
\end{table}

\section{Conclusions}
\label{conclusions}

This research proposes an innovative metho\-do\-logy for concurrent horizontal and vertical federated learning with non-IID data called square federated learning, using FCMs as the model to train. This proposal takes advantage of the benefits of this new artificial intelligence paradigm that allows the sharing of private data in a secure way to train a sophisticated machine learning model even when the participants do not have the same samples and features space. 

In the experiments, it is shown that this new approach improves the mean accuracy of an initial FCM model in all four aggregations. However, the performance of the mean F1 score depends heavily on the aggregation method, with the constant weights version being the one where the mean F1 score improves in most cases after the federated learning process, all but 2 out of 22. After that, the accuracy-based federation increases the mean F1 score in all but 3 out of 22 cases, and then precision-based weights with 4 out of 22, due to the fact that the precision metric is optimised by the process and it is related to the F1 score. Nevertheless, the AUC-based aggregation methods show worse results, with the mean F1 score improving for less than 70\% of the datasets. 

The experiments' results show that square federated lear\-ning is a novel and effective way to solve horizontal and vertical federated learning in datasets where both the samples and the features spaces are different. 

The limitation of the study is mainly related to the dependency on FCMs. While this research demonstrates the benefits and versatility of FCMs, this approach requires further investigation to be used with other techniques. Despite this limitation, it is important to highlight that this research remains relevant and represents a significant first step in the direction of com\-bi\-ning vertical and horizontal federated learning. This research lays a solid foundation for future research aimed at expanding the applicability of this approach in federated learning.

\section{Acknowledgements}
Prof. Salmeron research was kindly supported by the project Artificial Intelligence for Healthy Aging (Convocatoria 2021 – Misiones de I+D en Inteligencia Artificial: Inteligencia Artificial distribuida para el diagnóstico y tratamiento temprano de enfermedades con gran prevalencia en el envejecimiento, exp.: MIA.2021.M02.0007) led by Capgemini.

\Urlmuskip=0mu plus 1mu\relax
\bibliographystyle{elsarticle-num} 
\bibliography{sqfcm}

\end{document}